\documentclass[a4paper]{article}[11pt]
\usepackage{times,amsmath}
\usepackage{amssymb}
\usepackage{algorithm2e}
\usepackage{amsfonts}
\usepackage{graphicx}
\usepackage{graphics}
\usepackage{mathtools}
\usepackage{algpseudocode}
\usepackage{amsmath}
\usepackage{textcomp}
\usepackage{wrapfig,lipsum,booktabs}
\usepackage{multirow}
\usepackage[affil-it]{authblk}
\usepackage{fancyhdr}
\usepackage[named]{algo}

\begin{document}
\title{Mining  Recurrent Concepts in Data Streams using the Discrete Fourier Transform}

\author{Sakthithasan Sripirakas and Russel Pears}
\affil{Auckland University of Technology}
\maketitle

\begin{abstract}
In this research we address the problem of capturing recurring concepts in a data stream environment. Recurrence capture enables the re-use of previously learned classifiers without the need for re-learning while providing for better accuracy during the concept recurrence interval. We capture concepts by applying the Discrete Fourier Transform (DFT) to Decision Tree classifiers to obtain highly compressed versions of the trees at concept drift points in the stream and store such trees in a repository for future use. Our empirical results on real world  and synthetic data exhibiting varying degrees of recurrence show that the Fourier compressed trees are more robust to noise  and are able to capture recurring concepts  with higher precision than a meta learning approach that chooses to re-use classifiers in their originally occurring form.

\end{abstract}

\section{Introduction}
Data stream mining has been the subject of extensive research over the last decade or so.  One of the major issues with data stream mining is dealing with concept drift that causes models built by classifiers to degrade in accuracy over a period of time. 


While data steam environments require that models are updated to reflect current concepts, the capture and storage of recurrent concepts allows a classifier to use an older version of the model that provides a better fit with newly arriving data in place of the current model. This approach removes the need to explicitly re-learn the model, thus improving both accuracy and computational cost. A number of methods have been proposed that deal with the capture and exploitation of recurring concepts \cite{joa:trc}, \cite{gom:trc}, \cite{hos:nmo}, \cite{ali:jit} and  \cite{mor:anh}. Although achieving higher accuracy as expected during phases of concept recurrence in the stream, a major issue with existing approaches is the setting of user defined parameters to determine whether a current concept matches with one from the past.

Such parameters are difficult to set, particularly due to the drifting nature of real world data streams. Our approach avoids this problem by applying the Discrete Fourier Transform (DFT) as a meta learner. The DFT, when applied on a concept (Decision Tree model) results in a spectral representation that captures the classification power of the original models. One very attractive property of the Fourier representation of  Decision Tree is that most of the energy and classification power is contained within the low order coefficients \cite{hhil:afs}. The implication of this is that that when a concept C recurs as concept C* with relatively small differences caused by noise or concept drift, then such differences are likely to manifest in the high order coefficients of spectra S and S* (derived from C and C* respectively), thus increasing the likelihood of C* being recognized as a recurrence of C.

The DFT, apart from its use in meta learning, has a number of other desirable properties that make it attractive for mining high speed data streams. This includes the ability to classify directly from the spectra generated, thus eliminating the need for expensive traversal of a tree structure. 

Our experimental results in section 5 clearly show the accuracy, processing speed and memory advantages of applying the DFT as opposed to the meta learning approach proposed by Gama and Kosina in \cite{joa:trc}.
 
The rest of the paper is as follows. In section 2 we review work done in the area of capturing recurrences. We describe the basics of applying the DFT to decision trees in section 3. In section 4 we discuss a novel approach to optimizing the computation of the Fourier spectrum from a Decision Tree. Our experimental results are presented in section 5 and we conclude the paper in section 6 where we draw conclusions on the research and discuss some directions for future research.

\section{Related Research}
While a vast literature on concept drift detection exists \cite{pea:dci} only a small body of work exists so far on exploitation of recurrent concepts. The methods that exist fall into two broad categories. Firstly, methods that store past concepts as models and then use a meta learning mechanism to find the best match when a concept drift is triggered \cite{joa:trc}, \cite{gom:trc}. Secondly, methods that store past concepts as an ensemble of classifiers.  

Lazarescu in \cite{mih:aml} proposes an evidence forgetting mechanism for data instances based on a multiple window approach and a prediction module to adapt classifiers based on an estimation of the future rate of change. Whenever the  difference between the observed and estimated rates of change are above a user defined threshold a classifier that best represents the current concept is stored in a repository. Experimentation on the STAGGER dataset showed that the proposed approach outperformed the FLORA method on classification accuracy with re-emergence of previous concepts in the stream. 

Ramamurthy and Bhatnagar \cite{sas:trc} use an ensemble approach based on a set of classifiers in a global set G. An ensemble of classifiers is built dynamically from a collection of classifiers in G if none of the existing individual classifiers are able to meet a minimum accuracy threshold based on a user defined acceptance factor. Whenever the ensemble accuracy falls below the accuracy threshold, then the global set G is updated with a new classifier trained on the current chunk of data.

Another ensemble based approach by Katakis et al. is proposed in \cite{ioa:aeo}. A mapping function is applied on data stream instances to form conceptual vectors which are then grouped together into a set of clusters. A classifier is incrementally built on each cluster and an ensemble is formed based on the set of classifiers. Experimentation on the Usenet dataset showed that the ensemble approach produced better accuracy than a simple incremental version of the Naive Bayes classifier. 

Gomes et al  \cite{gom:trc} used a two layer approach with the first layer consisting of a set of classifiers trained on the current concept while the second  contains  classifiers created from past concepts. A concept drift detector is used to flag changes in concept and when a warning state is triggered incoming data instances are buffered in a  window to prepare a new classifier. If the number of instances in the warning window is below a user defined threshold then the classifier in layer 1 is used instead of re-using classifiers in layer 2. One major issue with this method is validity of the assumption that explicit contextual information is available in the data stream.

Gama and Kosina also proposed a two layered system in \cite{joa:trc} designed for delayed labeling, similar in some respects to the Gomes et al. \cite{gom:trc} approach. In their approach Gama and Kosina pair a base classifier in the first layer with a referee in second layer. Referees learn regions of feature space which its corresponding base classifier predicts accurately and is thus able to express a level of confidence on its base classifier with respect to a newly generated concept. The base classifier which receives the highest confidence score is selected, provided that it is above a user defined hit ratio parameter; if not, a new classifier is learned.

\section{Application of the Discrete Fourier Transform on Decision Trees}
The Discrete Fourier Transform (DFT) has a vast area of application in very diverse domains such as time series analysis, signal processing, image processing and so on. It turns out as Park \cite{par:kdf} and Kargupta \cite{hhil:afs} show that the DFT is very effective in terms of classification when applied on a decision tree model. Kargupta and Park in \cite{hhil:afs} explored the use of the DFT in a distributed environment but did not explore its usage in a data stream environment as this research sets out to do.

\begin{wrapfigure}{r}{0.2\textwidth}
\includegraphics[width=0.2\textwidth]{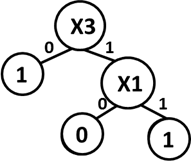}
\caption{Decision Tree with 3 binary features}
\label{fig:tree}
\end{wrapfigure}

Kargupta and Park in \cite{hhil:afs}  showed that the Fourier spectrum consisting of a set of Fourier coefficients fully captures a decision tree in algebraic form, meaning that the Fourier representation preserves the same classification power as the original decision tree.

A decision tree can be represented in compact algebraic form by applying the DFT to the paths of the tree.  We illustrate the process by considering a binary tree for simplicity but in practice the DFT can be applied to non binary trees as well \cite{hhil:afs}. For  trees with a total of d binary valued features the  $j^{th}$  Fourier coefficient $\omega_j$ is given by:
\begin{equation}
\omega_j=\frac{1}{2^d}{\sum_x{f(x)}\psi_j(x)}
\end{equation}
where $f(x)$ is the classification outcome of path vector x and $\psi_j(x)$, the Fourier basis function is given by:
\begin{equation}
\psi_j(x)=(-1)^{(j.x)}
\end{equation}

Figure 1 shows a simple example with 3 binary valued features $x_1$, $x_2$ and $x_3$, out of which only $x_1$ and  $x_3$ are actually used in the classification.

As shown in \cite{par:kdf}  only coefficients for paths that are defined by attributes that actually appear in the tree need to be computed as all other coefficients are guaranteed to be zero in value. Thus any coefficient of the form $\omega_{*1*}$ will be zero since attribute $x_2$ does not appear in the tree. 

With the wild card operator * in place we can use equations (1) and (2) to calculate non zero coefficients. Thus for example we can compute: 
\begin{align*}
\omega_{000}&=\frac{4}{8}f(**0)\psi_{000}(**0)+\frac{2}{8}f(0*1)\psi_{000}(0*1)+\frac{2}{8}f(1*1)\psi_{000}f(1*1)=\frac{3}{4}\\
\omega_{001}&=\frac{4}{8}f(**0)\psi_{001}(**0)+\frac{2}{8}f(0*1)\psi_{001}(0*1)+\frac{2}{8}f(1*1)\psi_{001}f(1*1)=\frac{1}{4}
\end{align*} and so on.
In addition to the properties discussed above, the Fourier spectrum of a given decision tree has two very useful properties that make it attractive as a tree compression technique \cite{hhil:afs}:
\begin{enumerate}
 \item
\begin{description}
All coefficients corresponding to partitions not defined in the tree are zero.
\end{description}
\item
\begin{description}
The magnitudes of the Fourier coefficients decrease exponentially with their order, where the order is taken as the number of defining attributes in the partition.
\end{description}
\end{enumerate}
Taken together these properties mean that the spectrum of a decision tree can be approximated by computing only a small number of low order coefficients, thus reducing storage overhead. With a suitable thresholding scheme in place, the Fourier spectrum consisting of the set of low order coefficients is thus an ideal mechanism for capturing past concepts. 

Furthermore, classification of unlabeled data instances can be done directly in the Fourier domain as it is well known that the inverse of the DFT defined in expression (3) can be used to recover the classification vector, instead of the use of a tree traversal which can be expensive in the case of deep trees. Expression 3 uses the complex conjugate $\overline\psi_j(x)$ function for the inverse operation in place of the original basis function of $\psi_j(x)$.
\begin{equation}
f(x)=\sum_j{\omega_j\overline\psi_j(x)}
\end{equation}
Due to thresholding and loss of some high order coefficient values the classification value $f(x)$ for a given data instance x may need to be rounded to the nearest integer in order to assign the class value. For example, with binary classes a value for f is rounded up to 1 if it is in the range $[0.5,1)$ and rounded down to 0 in the range $(0,0.5)$.
\section{Exploitation of the Fourier Transform for Recurrent Concept Capture} 
We first present the basic algorithm used in section 5.1 and then go on to discuss an optimization that we used for energy thresholding in section 5.2.
\subsection{The FCT algorithm}
We use CBDT \cite{hoe:acb} as the base algorithm which maintains a forest of trees. This forest of trees is dynamic in the sense that it can adapt to changing concepts at drift detection points. We thus define the memory consumed by this forest as \emph{active}. 

We integrate the basic CBDT algorithm with the ADWIN \cite{bit:lft} drift detector to signal concept drift. At the first concept drift point the best performing tree (in terms of accuracy) is identified and the DFT is applied after energy thresholding after which the resulting spectrum is stored in the repository for future use if the current concept recurs. The spectra stored in the repository are  fixed in nature as the intention is to capture past concepts. At each subsequent drift point a winner model is chosen by polling both the active memory and the repository. If the winner emerges from the active memory, two checks are made before the DFT is applied. First of all, we check whether the difference in accuracy between the winner tree in active memory (T) and the best performing model in the repository is greater than a tie threshold $\tau$. If this check is passed then the DFT is applied to T and a further check is made to ensure that its Fourier representation is not already in the Repository. If the winner model at a drift point emerges from an already existing spectrum in the Repository then no Fourier conversion is applied on any of the trees in active memory. Whichever model is chosen as the winner it is applied to classify all unlabeled data instances until a new winner emerges at a subsequent drift point. The least performing model M having the lowest weighted accuracy function is deleted if the repository has no room for new models.The weighted accuracy of M is defined by: $weight(M)=winner\_tally(M) * accuracy(M)$, where $winner\_tally$ is the number of times that M was declared a winner since it was inserted into the repository.

\begin {figure} [h!]
\setlength{\abovecaptionskip}{-2pt}
\setlength{\belowcaptionskip}{-20pt}
\begin{algorithm}{FCT}{
\label{cbdt}	
\qinput{Energy Threshold {\it $E_T$} , Accuracy Tie Threshold ${\tau}$}
\qoutput{Best Performing model M that suits current concept}}
read an instance I from the data stream \\
\qrepeat \\
	if  best  model M is in repository call Classify to classify I\\
	append 0 to ADWIN's window for M if classification is correct, else append 1
\quntil {drift is detected by ADWIN}\\
\qif {M is from active memory}\\
	identify best performing model F in repository \\
	\qif{(accuracy(M)-accuracy(F))$>\tau$} \\
		apply DFT on model M to produce F* using energy threshold {\it $E_T$} \\
		\qif{F* is not already in repository}\\
			insert F* into repository  \qfi \qfi \qfi \\
Identify best performing model M by polling active memory and repository \\
GoTo  1
\end{algorithm}
\end{figure}

\begin {figure} [h!]
\setlength{\abovecaptionskip}{-2pt}
\setlength{\belowcaptionskip}{-20pt}
\begin{algorithm}{Classify}{
\label{classifyusingfourierspectrum}	
\qinput{Instance {$I$}, Classifier ${M}$}
\qoutput{class value}}
\qif {M is a Decision Tree, route I to a leaf and return the class label of the leaf}\\
\qelse {using all coefficients ($j$) of ${M}$, Calculate $f(x)$ using $f(x)=\sum_j{\omega_j\overline\psi_j(x)}$ where  $\overline\psi_j(x)$ is the the complex conjugate function of $\psi_j(x)$ and $x$ is the instance $I$}\\
If $f(x)$ is greater than 0.5, return class1, otherwise class2
\qfi
\end{algorithm}
\end{figure}
\subsection{Optimizing the Energy Thresholding Process}
In order to avoid unnecessary computation of higher order coefficients which yield increasingly low returns on classification accuracy, energy threshold is highly desirable. To threshold on energy a subset S of the  (lower order) coefficients needs to be determined such that $\frac{E(S)}{E(T)}>\epsilon$, where $E(T)$ denotes the total energy across the spectrum and $\epsilon$ is the desired energy threshold value. 

In our optimized thresholding, we first compute the  cumulative energy $CE_i$ at order i given by:
$CE_i=\sum_{j=0}^i\sum_{k}({w_k}^2|order(k)=j)$. 

Given an order i, an upper bound estimate for the cumulative energy across the rest of the spectrum is given by: $(d+1-(i+1)+1)CE_i$, as the exponential decay property ensures that the energy at each of the orders $i+1$, $i+2$, $\cdots$, $d$ is less than energy $E_i$ at order i, where d is number of attributes in the dataset. Thus a lower bound estimate for the  fraction of the cumulative energy $CEF_i$ at order i to the total energy across all orders can then be expressed as:
\begin{equation}
\label{equ:estimatedcumulativeenergy}
CEF_i=\frac{CE_i}{CE_i+(d-i+1)E_i}
\end{equation} where $E_i$ is actual (computed) energy at order i. The lower bound estimate allows the specification of a threshold $\epsilon$ based on the energy captured by a given order i which is more meaningful to set rather than an arbitrary threshold.

The scheme expressed by equation (\ref{equ:estimatedcumulativeenergy}) enables the thresholding process to be done algorithmically. If the cumulative energy $CEF_i \ge \epsilon$, then we can guarantee that the actual energy captured is at least $\epsilon$, since $CEF_i$ is a lower bound estimate. On the other hand if $CEF_i < \epsilon$, then $CEF_{i+1}$ can be expressed as:
\begin{align}
\label{euqation2}
CEF_{i+1}&=\frac{CE_{i+1}}{CE_{i+1}+(d-i)E_{i+1}}=\frac{CE_i+E_{i+1}}{CE_i+dE_{i+1}}
\end{align}

Thus equation (\ref{euqation2}) enables the cumulative fraction to be easily updated incrementally for the next higher order (i+1) by simply computing the energy at that order while exploiting the exponential decay property of Fourier spectrum. The thresholding method guarantees that no early termination will take place. This is because $CEF_i$ is a lower bound estimate and hence the order that it returns will never be less than the true order that captures a given fraction $\epsilon$ of the total actual energy in the spectrum.
\section{Experimental Study}
This section elaborates on our empirical study involving the following learning systems: CBDT, FCT (Fourier Concept Trees) and MetaCT. The FCT incorporates the Fourier compressed trees in a repository in addition to the forest of trees that standard CBDT maintains. We  implement Gama's meta learning approach with CBDT as the base learner, namely MetaCT.  The main focus of the study is to assess the extent to which recurrences are recognized using old models preserved in classifier pools. 
\subsection{Parameter Values}
All experimentation was done with the following parameter values:
\begin{itemize}
    \item Hoeffding Tree Parameters
                The desired probability of choosing the correct split attribute=0.99, Tie Threshold=0.01, Growth check interval=32
     \item Tree Forest Parameters
                Maximum Node Count=5000, Maximum Number of Fourier Trees=50, Accuracy Tie Threshold $\tau$=0.01
     \item ADWIN Parameters
            drift significance value=0.01, warning significance value=0.3 (MetaCT only)
\end{itemize}
All experiments were done on the same software with C\# .net runtime and hardware with Intel i5 CPU and 8GB RAM, clearning the memory in each run to have a fair comparison.
\subsection{Datasets used for the experimental study}
\label{sec:datasets}
We experimented with data generated from 3 data generators commonly used in drift detection and recurrent concept mining, namely SEA concept \cite{str:ase}, RBF  and Rotating hyperplane generators. In addition we used 2 real-world datasets,  {\it Spam } and the {\it NSW electricity} which have also been commonly used in previous research. 

For the synthetic datasets, each of the 4 concepts spanned 5,000 instances and reappeared 25 times in a data set, yielding a total of 500,000 instances with 100 true concept drift points. 

In order to challenge the concept recognition process, we added a 10\% noise level for all synthetic data sets to ensure that concepts recur in similar, but not exact form. 
 
\subsubsection{Synthetic Data Sets}
We used MOA \cite{bit:moa} as the tool to generate these datasets.
\begin{enumerate}
    \item {\bf SEA: }
The concepts are defined by the function $feature1 + feature2 > threshold$.  We ordered the concepts as concept1, concept2, concept3 and concept4 generated using threshold values 8,7,9 and 9.5 respectively on the first data segment of size 20,000. We generated 96 recurrences of a modified form of these concepts by using different seed values in MOA for each sequence of recurrence. Thus, our version of this dataset differed from the one used by Gama and Kosina \cite{joa:trc}. who simply used 3 concepts with the third being an exact copy of the first.

	 \item {\bf RBF: }
The number of centroids parameter was adjusted to generate different concepts for the RBF dataset. Concept1, concept2, concept3 and concept4 were produced with the number of centroids set to 5,15, 25 and 35 respectively. Similar to the SEA dataset, the seed parameter helped in producing similar concepts for a given centroid count value. 
This dataset had 10  attributes.
	 \item {\bf Rotating hyperplane: }
The number of drifting attributes was adjusted to 2,4,6, and 8 in a 10 dimensional data set to create the four concepts. The concept ordering, generation of similar concepts and concatenation were exactly the same as in the other data sets mentioned above.
\end{enumerate}
\subsubsection{Real World datasets}
\begin{enumerate}
    \item {\it Spam Data Set: } The Spam dataset was used in it original form \footnote{from {\it http://www.liaad.up.pt/kdus/products/datasets-for-concept-drift}} which encapsulates  an evolution of Spam messages. There are 9,324 instances and 499 informative attributes, which was different from the one version used by Gama that had 850 attributes. 
	 \item {\it Electricity Data Set: }
NSW Electricity dataset is also used in its original form \footnote{from {\it http://moa.cms.waikato.ac.nz/datasets/}}. There are two classes {\it Up} and {\it Down} that indicate the change of price with respect to the moving average of the prices in last 24 hours.
\end{enumerate}

\subsection{Tuning MetaCT Key Parameters}
In our preliminary experiments, we found optimal values for the two parameters, {\it delay} in receiving labels for the instances in short term memory, and {\it hit percentage threshold value} as 200 and 80\%, respectively. The latter parameter reflects the estimated similarity of the current concept with one from the past and thus controls the degree of usage of classifiers from the pool.
\subsection{Comparative study: CBDT vs FCT vs MetaCT}
Our focus in this series of experiments was to assess the models in terms of accuracy, memory consumption and processing times. None of the previous studies reported in the  recurrent concept mining literature undertook a comparative study against other approaches and so we believe our empirical study to be the first such effort. Furthermore, all of the previous studies focused exclusively on accuracy without tracking memory and execution time overheads and so this study would also be the first of its kind.
\subsubsection{Accuracy}
A delay period of 200 was used with all three approaches in order to perform a fair comparison. Figure 2 clearly shows that overall, FCT significantly outperforms its two rivals with respect to classification accuracy. The major reason for FCT's superior performance was its ability to re-use previous classifiers as shown in the segment 20k-25k on the RBF dataset where the concept is similar to concept1 that occurred in interval 1-5K. This is in contrast to MetaCT which was unable to recognize the recurrence of concept1. A similar situation occurs in the interval 25k-35k where the concept  is similar to the previously occurring concepts, which are concept2 and 3. As expected CBDT, operating on its own without support for concept recurrence had a relatively flat trajectory throughout the stream segment.


A similar trend to the RBF dataset was observed in Rotating Hyperplane and SEA datasets as well. It can be clearly seen that FCT was successful in reusing the models learned before on data segments from 20k to 25k and from 30k to 35k. Though a preserved model was reused on the data segment from 25k to 30k (corresponding to concept3), the accuracy was not as high as in the above two segments. On the segment from 35k to 40k, concept recurrence was not picked up by either FCT or MetaCT resulting in a new classifier being used.

We omit the figure for the SEA dataset due to space constraints. In summary, FCT outperformed MetaCT over 90 recurring concepts whereas MetaCT did better in 6 occurrences, thus maintaining the same trend as with the other 2 synthetic datasets that we experimented with.

\begin{wrapfigure}{r}{0.7\textwidth}
  \begin{center}
    \includegraphics[width=0.7\textwidth]{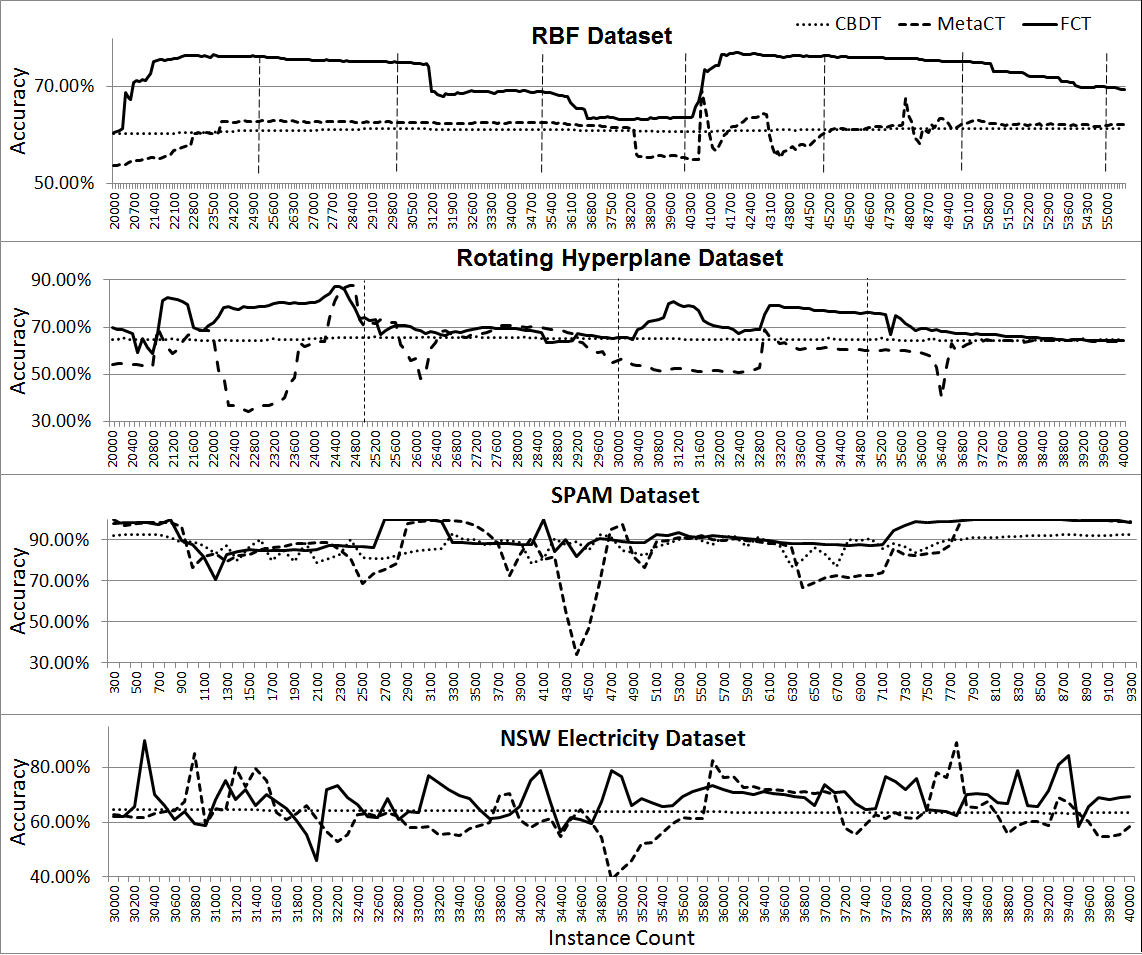}
  \end{center}
  \caption{Classification Accuracy for CBDT, FCT and MetaCT}
\label{accuracycurves}
\end{wrapfigure}

The next experiment was  on the NSW Electricity data set. Figure \ref{accuracycurves} shows that overall, FCT was the winner here as well, outperforming MetaCT at 25 segments out of 35 that we tracked. Sudden fall in accuracy of MetaCT is occational but due to incorrect selection of winner which was a decision stump.

\subsubsection{Memory}
Our experimentation on accuracy has revealed, especially in the case of FCT, the key role that concept capture and re-use has played in improving accuracy. The question is, what price has to be paid in terms of memory overhead in storing these recurrent concepts? Table1 clearly shows that the Fourier transformed trees consume a small fraction of the memory used by the pool of trees kept in FCT's active memory, despite the fact that collectively these models outperform their decision tree counterparts at a greater number of points in the stream progression.

Comparison of overall memory consumption across FCT and MetaCT is complicated by the fact that the latter tended to have immature trees in its classifier pool that under fits concepts. Despite this, Table 1 reveals that FCT's memory consumption is competitive to that of MetaCT. The only case where MetaCT had a substantially lower consumption was with the Spam dataset with a lower overhead for active memory. 
\subsubsection{Processing Speed}
FCT and MetaCT have two very contrasting methods of classification. The former routes each unlabeled instance to a single tree, which is the best performing tree selected at the last concept drift point. In contrast MetaCT classifies by routing an unlabeled instance to all referees to obtain their assessment of their corresponding models and in general will have more processing overhead on a per instance basis. However, FCT has potentially more overhead at concept drift points if the winner tree is one that is selected from the active forest as this tree needs to be converted into its Fourier representation. Thus it is interesting to contrast the run time performances of the two approaches. 

\begin{table}
\scriptsize
\label{tab:overheadcomparison}
\caption{Average Memory Consumption (in KBs) and Processing Speed Instances per second) Comparison}
  \begin{tabular}{| l | l | l | l | l | l | l |}
    \hline
    \multirow{4}{*}{Datasets} & \multicolumn{2}{c|}{ Memory } & \multicolumn{2}{c|}{ Memory}  & {Processing} & {Processing}\\
	          &  \multicolumn{2}{c|}{FCT} & \multicolumn{2}{c|}{MetaCT} & {Speed} & {Speed} \\
     & Tree & Fourier & Tree & Pool & & \\ 
	  & Forest & Pool & Forest & & FCT & MetaCT \\ \hline
    RBF & 97.9 & 24.8  & 122.7  & 14.9 & 3540.6 &  2662.5\\ \hline
    Rot. Hy/plane & 187.4   & 59.7  & 148.7  & 43.4 & 2686.2 & 2180.1\\ \hline
	 SEA & 29.3  & 34.8  & 28.0  & 18.1 & 11368.2 & 10125.8 \\ \hline
    Spam & 1712.8  & 18.8  & 878.0 & 15.3  & 4.1 & 4.3 \\ \hline
    Electricity & 48.4  & 39.9  & 19.8  & 18.9  & 5705.7 & 7191.42 \\ \hline    
  \end{tabular}
\end{table}

Table 1 shows that in general FCT has a higher processing speed (measured in instances processed per second); the only exception was with the Electricity dataset where MetaCT was faster. The electricity data contains a relatively larger number of drift points in comparison to the other datasets and this in turn required a greater number of DFT operations to be performed, thus slowing down the processing. In our future research we will investigate methods of optimizing the DFT process.

Finally, we close this section with two general observations on FCT which hold across all experiments reported above. Firstly, we note that the Discrete Fourier Transform (DFT), as expected, was able to capture the \emph{essence} of a concept to the extent that when it reappeared in a modified form in the presence of noise it was still recognizable and was able to classify it accurately. Secondly, not only was the DFT robust to noise, it actually performed better than the original decision trees at concept recurrence points due to its better generalization capability.

\subsection{Sensitivity Analysis  on FCT}
Having established the superiority of FCT we were interested in exploring the sensitivity of FCT's accuracy on two key factors.

\begin{wrapfigure}{r}{0.5\textwidth}
	\centering
    \includegraphics[width=0.5\textwidth]{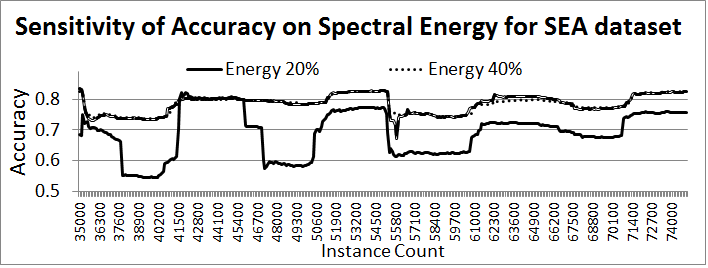}
  \caption{Sensitivity of Accuracy on Spectral Energy}
\label{energyaccuracy}
\end{wrapfigure}
\subsubsection{Energy Threshold}
FCT's energy threshold parameter controls the extent to which it captures recurring contexts. We ran experiments with all datasets we experimented with and tracked accuracy across four different thresholds: 95\%, 80\%, 40\% and 20\%. The trends observed for all datasets were very similar and hence we display results for the SEA concepts dataset due to space constraints. Figure \ref{energyaccuracy} clearly shows that very little difference in accuracy 
exists between the trajectories for 40\% and 95\%, showing the resilience of the DFT in capturing the classification power of concepts at low energy levels such as 40\%. Thus the low order Fourier coefficients that survive the 40\% threshold hold almost the same classification power of spectra at the 80\% or 90\% levels which contain more coefficients. Such higher energy spectra  would represent larger decision trees in which some of the decision nodes would be split into leaf nodes, thus enabling them to reach a slightly higher level of accuracy.

\subsubsection{Noise Level}
In section 5.4 we observed that FCT outperformed MetaCT by recognizing concepts from the past even though the concepts did not recur exactly in their original form due partly to noise and partly due to different data instances being produced as a result of re-seeding of the concept generation functions. In this experiment we explicitly test the resilience of FCT to noise level by subjecting it to three different levels of noise - 10\%, 20\% and 30\%. For reasons of completeness we also included MetaCT in he experimentation to aid in the interpretation of results. 

Figure 4 reveals three interesting pieces of information.  Firstly, FCT is still able to recognize recurring concepts at the 20\% noise level even though the models it re-uses do not have quite the same classification power (when compared to  the 10\% noise level) on the current concept due to data instances being corrupted by a relatively higher level of noise. 

Secondly, FCT's concept recurrence recognition is essentially disabled at the 30\% noise level as shown by its flat trajectory, thus essentially performing at the level of the base CBDT system. It is able to avoid drops in accuracy on account of the forest of trees that is maintained and is able to switch quickly and seamlessly from one tree to another when concepts change occurs.

Thirdly, although MetaCT is not the focus of this experiment we see that MetaCT's ability to recognize recurring concepts is disabled at the 20\% level, showing once gain the resilience of the DFT to noise. At the 30\% level its accuracy drops quite sharply at certain points in the stream.This is due to the fact that it learns a single new classifier and relies on it to classify instances in the current concept. In contrast, FCT exploits the entire forest of trees and switches from one tree to another tree in its active forest in response to drift.
\begin{figure}
\setlength{\abovecaptionskip}{-2pt}
\setlength{\belowcaptionskip}{-20pt}
\label{accuracynoise}
\centering
\includegraphics[scale=0.36]{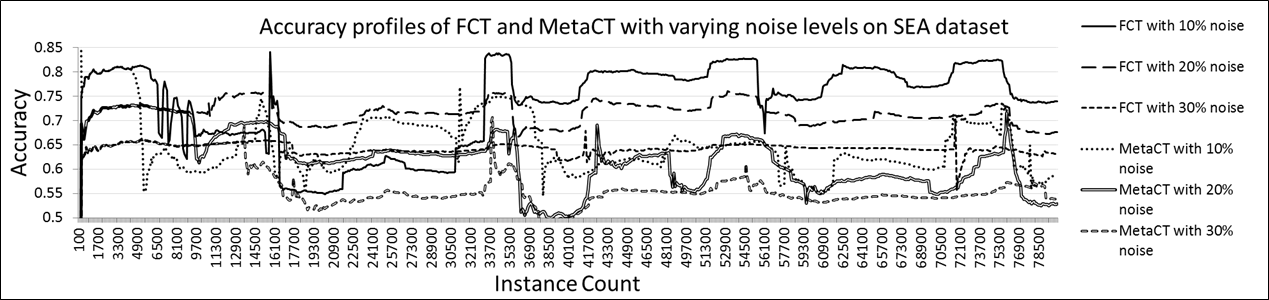}
\caption{Sensitivity of Accuracy for FCT and MetaCT on Noise Level}
\label{curves}
\end{figure}

\section{Conclusions and Future Work}
In this research we proposed a novel mechanism for mining data streams by capturing and exploiting recurring concepts. Our experimentation showed that the Discrete Fourier Transform when applied on Decision Trees captures concepts very effectively, both in terms of information content and conciseness. The Fourier transformed trees were robust to noise and were thus able to recognize concepts that reappeared in modified form, thus contributing significantly to improving accuracy. Overall our proposed approach significantly outperformed the meta learning approach by Gama and Kosina \cite{joa:trc} in terms of classification accuracy while being competitive in terms of memory and processing speed. 

We were able to optimize the derivation of the Fourier spectrum by an efficient thresholding process but there is further scope for optimization in the computation of low order coefficients in streams exhibiting frequent drifts, as our experimentation with the NSW Electricity dataset reveals. Our future work will concentrate on two areas. Firstly we plan to investigate the use of multi-threading on a parallel processor platform to optimize the DFT operation. Allocating the DFT process to a thread while another thread processes the incoming stream will greatly speed up processing for FCT as the two processes are independent of each other and can be executed in parallel. Secondly,  the computation of the Fourier basis function that requires a vector dot product computation can be optimized  by using patterns in the two vectors involved.


\end{document}